\documentclass[conference]{IEEEtran}
\IEEEoverridecommandlockouts

\usepackage{cite}
\usepackage{amsmath,amssymb,amsfonts}
\usepackage{algorithmic}
\usepackage{graphicx}
\usepackage{textcomp}
\usepackage[dvipsnames, table]{xcolor}
\usepackage{paralist}
\usepackage[inline]{enumitem}
\usepackage{caption}
\usepackage{makecell}
\usepackage{tabu, booktabs}
\usepackage{float}
\usepackage{subcaption}
\usepackage{comment}
\usepackage{todonotes}
\usepackage[breaklinks,colorlinks,bookmarks=true]{hyperref}
\usepackage{csquotes}
\usepackage{soul}
\usepackage{pifont}
\usepackage{multirow}
\usepackage{tcolorbox}
\usepackage{tabularx}
\usepackage{array}
\usepackage{colortbl}
\usepackage{siunitx}
\usepackage{fancyhdr}

\definecolor{cadetblue}{rgb}{0.37, 0.62, 0.63}
\definecolor[named]{xBlue}{HTML}{18647E}
\definecolor[named]{xOrange}{HTML}{FF9B00}
\definecolor[named]{xDarkBlue}{cmyk}{1,0.58,0,0.21}

\hypersetup{
    colorlinks=true,
    linkcolor=xOrange,
    filecolor=magenta,      
    urlcolor=xDarkBlue,
    citecolor=xBlue,
}

\usepackage[a4paper, total={184mm,239mm}]{geometry}
\def\BibTeX{{\rm B\kern-.05em{\sc i\kern-.025em b}\kern-.08em
    T\kern-.1667em\lower.7ex\hbox{E}\kern-.125emX}}
    
\ExplSyntaxOn
\DeclareExpandableDocumentCommand{\convertlen}{ O{cm} m }
 {
  \dim_to_decimal_in_unit:nn { #2 } { 1 #1 } cm
 }
\ExplSyntaxOff
\usepackage{gnuplottex}
\usepackage{tikz}
\usepackage{mathtools}

\usetikzlibrary{matrix,calc,positioning,arrows.meta}

\newcommand{\cmark}{\textcolor{green!70!black}{\ding{51}}}
\newcommand{\xmark}{\textcolor{red}{\ding{55}}}

\DeclareMathOperator*{\argmax}{arg\,max}

\newcommand*\circled[1]{\tikz[baseline=(char.base)]{\node[shape=circle,fill,inner sep=0.5pt] (char) {\textcolor{white}{#1}};}}

\usepackage[normalem]{ulem}

\begin{document}

\title{RankMap: Priority-Aware Multi-DNN Manager for Heterogeneous Embedded Devices}

\author{
\IEEEauthorblockN{Andreas Karatzas$^{1}$, Dimitrios Stamoulis$^2$, Iraklis Anagnostopoulos$^1$}
\IEEEauthorblockA{$^1$School of Electrical, Computer, and Biomedical Engineering, Southern Illinois University, Carbondale, IL, U.S.A.}
\IEEEauthorblockA{$^2$Department of Electrical and Computer Engineering, The University of Texas at Austin, Austin, TX, U.S.A.}
\IEEEauthorblockA{Email: \{andreas.karatzas, iraklis.anagno\}@siu.edu, dstamoulis@utexas.edu}
}

\maketitle

\thispagestyle{fancy}
\fancyhead{}
\rhead{}
\lhead{}
\chead{{\small Accepted for publication at the 28th Design Automation and Test in Europe Conference (DATE 2025)}}
\renewcommand{\headrulewidth}{0.4pt}

\begin{abstract}

Modern edge data centers simultaneously handle multiple Deep Neural Networks (DNNs), leading to significant challenges in workload management. Thus, current management systems must leverage the architectural heterogeneity of new embedded systems to efficiently handle multi-DNN workloads. This paper introduces RankMap, a priority-aware manager specifically designed for multi-DNN tasks on heterogeneous embedded devices. RankMap addresses the extensive solution space of multi-DNN mapping through stochastic space exploration combined with a performance estimator. Experimental results show that RankMap achieves $\times 3.6$ higher average throughput compared to existing methods, while preventing DNN starvation under heavy workloads and improving the prioritization of specified DNNs by $\times 57.5$.

\end{abstract}


\setlist{nosep}

\section{Introduction}

We are in an era of rapid machine learning advancements, where Deep Neural Networks (DNNs) play an important role in modern embedded systems. Embedded systems often utilize heterogeneous computing components to manage the computational demands of DNNs. However, run-time managers neglect the underlying heterogeneity~\cite{wu2019machine, karatzas2024mapformer}. Current deep learning frameworks typically utilize the CPU or the GPU exclusively, thus underutilizing the system's diverse components~\cite{hsieh2019surf, karatzas2024balancing}. 

Additionally, modern embedded devices must address the challenge of deploying multi-DNN workloads, where multiple DNNs run simultaneously on a single system. Efficiently managing the available computing resources in these scenarios becomes a critical issue. A way to solve this problem is to map each DNN on one of the available processing units~\cite{kang2020scheduling}. However, these coarse-grained methods are far from optimal~\cite{karatzas2023omniboost}.

Moreover, the prioritization of each DNN within a multi-DNN workload, an often overlooked aspect, adds significant complexity to the problem. Meeting Service Level Agreement (SLA) requirements become particularly challenging, making prioritization a crucial metric for evaluating a multi-DNN manager~\cite{wang2023cd}. Users are categorized into different SLA groups, leading to multi-DNN workloads where each DNN has a different priority level. For instance, in edge data centers where multiple users submit DNN queries, \textit{current runtime managers lead to reduced throughput in $91\%$ of cases and starving some DNNs $30\%$ of the time, resulting in an overall unsatisfactory user experience} (see Section~\ref{sec:motivation}).

In this work, we present \textbf{RankMap}, a framework for efficient multi-DNN management onto heterogeneous embedded devices. RankMap learns the computational profile of the individual DNN layers and groups them into pipeline stages to map them among the given computing components. It reduces resource contention and boosts system throughput. RankMap also ensures that each DNN receives enough computing resources proportional to its priority (rank) without starving other DNNs running concurrently. RankMap outperforms previous approaches, achieving up to $\times 3.6$ higher average throughput across all multi-DNN scenarios, while satisfying priority constraints $\times 57.5$ more efficiently, ensuring no DNN is starved.

\textbf{Overall, our main contributions are:}
\begin{inparaenum}
    \item[\circled{1}] We propose RankMap, a lightweight and scalable multi-DNN manager that utilizes fine-grained DNN segmentation to boost system throughput on heterogeneous embedded devices.
    \item[\circled{2}] RankMap creates mappings for multi-DNN workloads considering each DNN's priority to satisfy their performance requirements, avoiding starvation effects.
    \item[\circled{3}] RankMap employs a novel multi-task attention-based CNN for throughput estimation of any multi-DNN scenario. A Monte Carlo Tree Search (MCTS) algorithm utilizes this estimator as a feedback mechanism to efficiently explore the vast solution space of those mappings.
\end{inparaenum}

\section{Motivation}\label{sec:motivation}

In this section, we highlight the importance of an efficient multi-DNN manager. Specifically, we demonstrate \circled{1} the impact of DNN partitioning on system throughput, and \circled{2} potential starvation issues in multi-DNN workloads. We target scenarios where each DNN is an independent task, meaning the output of one DNN does not serve as input for another. This setup is particularly relevant for edge data centers serving multiple users, each submitting different DNN requests. We focus on Computer vision applications in our experiments; even with the rise of Large Language Models (LLMs) and Generative AI~\cite{paramanayakam2024lessmore}, vision models:
\begin{inparaenum}[(\bgroup\bfseries i\egroup)] 
    \item remain more prevalent in edge environments over natural language ones~\cite{chen2019deep, karatzas2024pythia}, 
    \item are typically executed on edge hardware more efficiently~\cite{huynh2017deepmon}, as 
    \item major hardware providers like ARM and NVIDIA offer more robust support for computer vision at the edge compared to natural language tasks~\cite{lin2023vila, armnn_blog}. 
\end{inparaenum}
To that end, we select four diverse and widely used DNNs for our multi-DNN workload: SqueezeNet-V2, Inception-V4, ResNet-50, and VGG-16. We utilize the Orange Pi 5 board, which features a Mali-G610 GPU and a big.LITTLE CPU, consisting of a quad-core Cortex-A76 at 2.4GHz and a quad-core Cortex-A55 at 1.8GHz. We introduce the potential throughput metric $\mathcal{P}$ to quantify a DNN's performance in a multi-DNN environment relative to its isolated performance, \textit{offering insights into resource utilization efficiency.} For a DNN $i$, $\mathcal{P}^i = \frac{t_{current}^{i}}{t_{ideal}^{i}}\label{eq:motiv}$, where $t_{current}^{i}$ is the throughput of DNN $i$ (in inferences per second) when running alongside other DNNs, and $t_{ideal}^{i}$ is its throughput when running alone on the GPU.

\begin{figure}[t]
    \centering
    \resizebox{1.0\columnwidth}{!}{\includegraphics[page=1, width=\linewidth, clip]{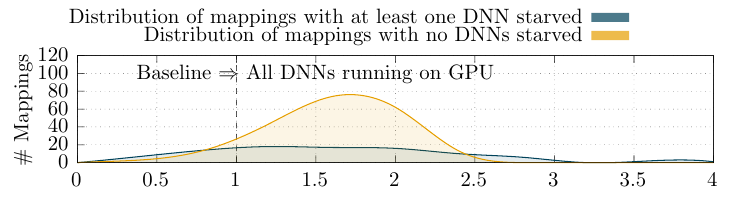}}
    \caption{Average throughput $\mathcal{T}$ normalized by baseline.}
    \label{fig:motiv:throughput}
\end{figure}

\begin{figure}[t]
    \centering
    \resizebox{1.0\columnwidth}{!}{\includegraphics[page=1, width=\linewidth, clip]{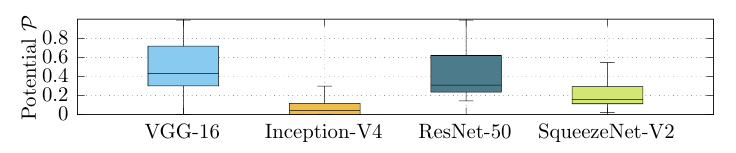}}
    \caption{Potential Throughput $\mathcal{P} \forall$ DNNs.}
    \label{fig:motiv:priority}
\end{figure}

First, we mapped all the DNNs onto the GPU, our baseline configuration, as it's traditionally preferred over the CPU for its superior computing capabilities. Next, we generated 300 unique random mappings of the selected multi-DNN workload. In these mappings, each DNN was split into arbitrary groups of contiguous layers, forming pipeline stages. These stages were then randomly assigned to the three available computing components—the big CPU cluster, the LITTLE CPU cluster, and the GPU—resulting in a distributed pipeline execution across the heterogeneous system.
Figure~\ref{fig:motiv:throughput} presents the throughput distribution normalized to the baseline. For each mapping, throughput is defined as $\mathcal{T} = \frac{\sum^{N}_{i=1} t_{current}^{i}}{N}$, where $N$ is the number of DNNs in the workload—four in our case. Figure~\ref{fig:motiv:priority} shows the distribution of the potential throughput $\mathcal{P}$ for each DNN. From these figures, we observe the following:

\begin{tcolorbox}[
  colback=cadetblue!5,
  colframe=cadetblue,
  colbacktitle=cadetblue!20,
  coltitle=black,
  title=Key Observations,
  fonttitle=\bfseries
]
\setlist{labelwidth=\widthof{0.},leftmargin={\labelwidth},rightmargin=0pt}
\begin{enumerate}
    \item[\circled{1}] The baseline heavily saturates the GPU, and hence $91 \%$ of mappings exhibit better average throughput $\mathcal{T}$. This proves the efficiency of partitioning DNNs across all the available computing components.
    \item[\circled{2}] A multi-DNN manager is likely to suboptimally allocate the DNNs across the computing components. This can significantly reduce the performance of at least one DNN and starve it. Specifically, in Figure~\ref{fig:motiv:throughput}, we observe that the risk of starvation is almost a certainty when we go beyond the threshold of $\mathcal{T} \ge 2.4$. Mappings with the highest throughput $\mathcal{T}$ often come at the expense of other DNNs in the workload, with $30.2 \%$ of mappings causing starvation.
    \item[\circled{3}] Deeper and more complex architectures, such as Inception-V4 and VGG-16, are very challenging due to their high probability of starvation. Specifically, in Figure~\ref{fig:motiv:priority}, we observe that the mean $\mathcal{P}$ for Inception-V4 is around $0.1$.
    \item[\circled{4}] Finding mappings that consistently satisfy minimum performance criteria is improbable. In Figure~\ref{fig:motiv:priority}, we observe that going beyond the threshold of $\mathcal{P} \ge 0.6$ means some DNNs will underperform.
    \item[\circled{5}] In Figure~\ref{fig:motiv:priority}, we observe that more than $60 \%$ of DNNs will exhibit $\mathcal{P} \le 0.2$. Thereby, to go beyond that barrier, we have to research AI-powered multi-DNN managers that efficiently correlate the hardware properties to the queried multi-DNN workloads.
\end{enumerate}
\end{tcolorbox}

Overall, supporting priorities for DNNs to \textit{ensure they meet performance goals while maintaining high system throughput and preventing application starvation} is a challenging and unexplored problem.

\section{Related Work}\label{sec:related}

The authors in~\cite{hsieh2019surf} propose SURF, which employs heuristic techniques for DNN partitioning to construct efficient DNN pipelines. Similarly, Pipe-it~\cite{wang2019high} is a framework that predicts the performance of different layers to create DNN pipelines. However, it targets the CPU only.
Furthermore, the authors in~\cite{aghapour2024arm} propose the ARM-CO-UP framework, which boosts throughput by employing pipelined execution of DNN partitions. However, they do not consider concurrent execution of different DNNs.
BAND~\cite{jeong2022band} focuses on identifying and grouping DNN sub-graphs that utilize similar computing operations. Moreover, the authors in~\cite{kwon2021heterogeneous} present a method that allocates different layers to the system's various processing units to prioritize immediate resource assignment. However, they follow a greedy approach, which is not scalable. 
MOHaM~\cite{das2024multi} co-optimizes hardware mapping for multi-DNN workloads. However, it designs sub-accelerators rather than focusing on DNN partitioning or concurrent DNN execution on shared resources.
Similarly, the authors in~\cite{spantidi2022targeting} present a heuristic method to support multi-DNN workloads. However, they employ multiple accelerators operating on different precisions that may not be available on conventional embedded devices. Furthermore, the authors in~\cite{archer2023pipeline} demonstrate a method to pipeline the parallelism of DNNs by partitioning them into smaller stages, thereby reducing inference latency. However, their method targets larger systems and does not support priorities.
Likewise, MOSAIC~\cite{han2019mosaic} uses DNN partitioning for workload distribution, relying on a linear regression model that correlates layer input sizes with computational needs, trained on single DNN cases. This method overloads the embedded GPU and cannot support DNNs with different priorities.
ODMDEF~\cite{lim2021odmdef} creates pipelines for handling multiple DNN workloads. This method, though, needs a considerable amount of data to achieve reliable accuracy and does not support priorities either.
The authors in~\cite{kang2020scheduling} introduce a framework based on an evolutionary algorithm to partition the DNNs on the given computing components. However, this method converges slowly and does not scale well due to the constant re-training required to support different scenarios. Additionally,
the authors in~\cite{karatzas2023omniboost} present OmniBoost, a framework that uses a CNN to estimate the performance of DNN layers on the computing components of the embedded device. Even though OmniBoost achieves high average throughput, it does not support priorities and requires extensive profiling of all the layers of a DNN. Our key differences from the state-of-the-art are manifold:
\begin{inparaenum}[(\bgroup\bfseries i\egroup)]
\item We efficiently explore the design space by employing MCTS for pruning a decision tree based on highly accurate feedback from our throughput estimator;
\item RankMap accounts for each DNN's priority to generate a mapping; and
\item We prevent DNN starvation regardless of the workload.
\end{inparaenum}
\begin{figure*}[!t]
    \centering
    \resizebox{1.0\textwidth}{!}{\includegraphics[page=1, width=\linewidth, clip, trim={2em 4 3 3}]{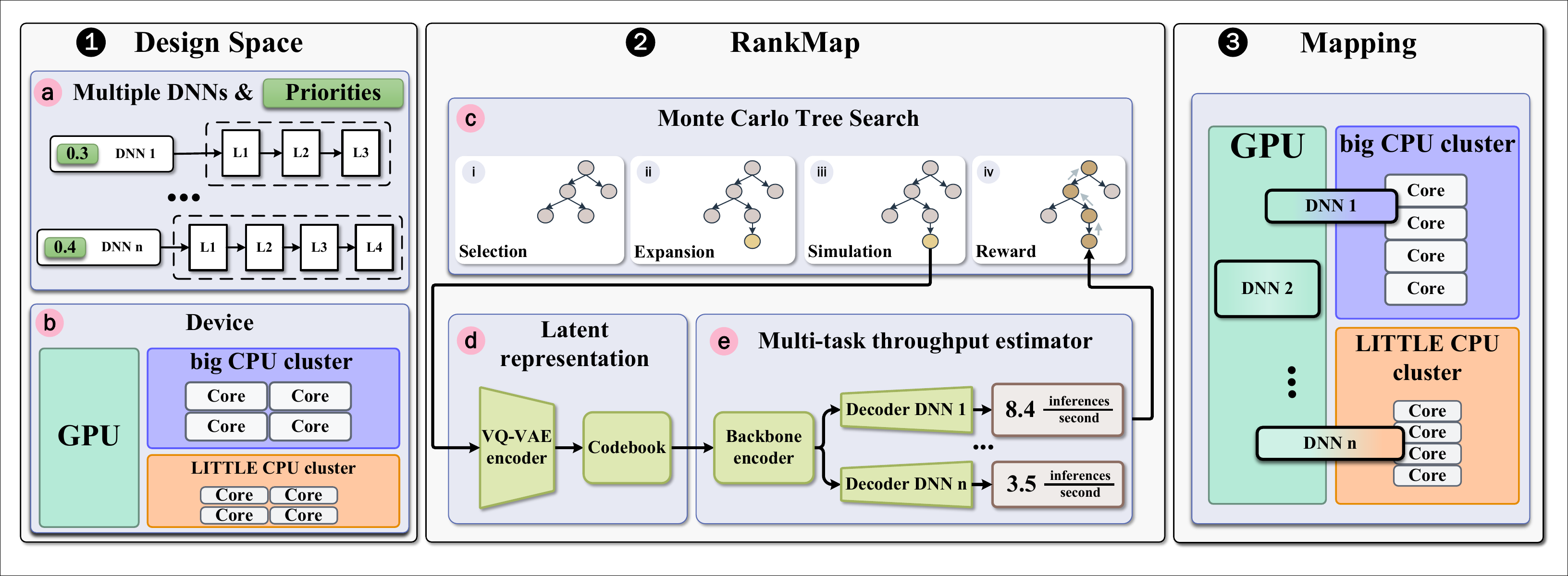}}
    \caption{A high-level overview of RankMap.}
    \label{fig:method:high-level-overview}
\end{figure*}
A qualitative comparison is presented in Table~\ref{tab:related_comparison}.
\begin{table}[h]
\setlength{\tabcolsep}{2pt}
\caption{Comparison between state-of-the-art and RankMap.}
\label{tab:related_comparison}
\footnotesize
\begin{tabularx}{\columnwidth}{
  >{\raggedright\arraybackslash}p{2.2cm}
  >{\centering\arraybackslash}p{1.23cm}
  >{\centering\arraybackslash}p{1.24cm}
  >{\centering\arraybackslash}p{0.79cm}
  >{\centering\arraybackslash}p{1.28cm}
  >{\centering\arraybackslash}p{1.39cm}
}
\toprule
\textbf{Feature} & \shortstack{MOSAIC\\\cite{han2019mosaic}} & \shortstack{ODMDEF\\\cite{lim2021odmdef}} & \shortstack{GA\\\cite{kang2020scheduling}} & \shortstack{OmniBoost\\\cite{karatzas2023omniboost}} & \cellcolor{green!20}\textbf{RankMap} \\
\midrule
Single-DNN$^1$ & \cmark & \cmark & \cmark & \cmark & \cellcolor{green!5}\cmark \\
\rowcolor{gray!5}
Multi-DNN$^2$ & \xmark & \xmark & \cmark & \cmark & \cellcolor{green!15}\cmark \\
DNN partitioning$^3$ & \cmark & \cmark & \cmark & \cmark & \cellcolor{green!5}\cmark \\
\rowcolor{gray!5}
High throughput$^4$ & \cmark & \cmark & \cmark & \cmark & \cellcolor{green!15}\cmark \\
Priority-aware$^5$ & \xmark & \xmark & \xmark & \xmark & \cellcolor{green!5}\cmark \\
\rowcolor{gray!5}
Fast training$^6$ & \xmark & \xmark & \xmark & \cmark & \cellcolor{green!15}\cmark \\
No starvation$^7$ & \xmark & \xmark & \xmark & \xmark & \cellcolor{green!5}\cmark \\
\bottomrule
\end{tabularx}
\raggedright\scriptsize
$^1$Predominantly supports management of one executing DNN. \\
$^2$Targets workloads with multiple concurrently executing DNNs. \\
$^3$Partitions DNNs into pipeline stages across computing components. \\
$^4$Optimizes average multi-DNN workload throughput. \\
$^5$Accounts for different DNN priorities in resource allocation. \\
$^6$Employs efficient training process with low training complexity. \\
$^7$Guarantees minimum resources for all DNNs in workload to prevent starvation.
\end{table}

\section{Proposed Framework}\label{sec:methodology}

This section introduces RankMap, a multi-DNN manager for heterogeneous embedded systems. Figure~\ref{fig:method:high-level-overview} depicts the overview of the proposed approach. RankMap takes three key inputs:
\begin{inparaenum}[(\bgroup\bfseries i\egroup)]
    \item A set of DNNs to be executed concurrently;
    \item A collection of available computing components, such as big CPU cluster, LITTLE CPU cluster, and GPU; and
    \item A list of application ranks indicating the priority of each DNN in the workload.
\end{inparaenum}
RankMap addresses the following objectives: \circled{1} Maintain high system throughput; \circled{2} Account for individual DNN priorities; \circled{3} Prevent application starvation. 

To this end, RankMap employs a fine-grained approach to workload distribution:
\begin{inparaenum}[(\bgroup\bfseries i\egroup)]
    \item \textbf{DNN partitioning:} Each DNN is divided into smaller sub-DNNs;
    \item \textbf{Adaptive Mapping:} These sub-DNNs are strategically allocated across the system's diverse computing components; and
    \item \textbf{Performance Optimization:} This granular distribution enhances the performance of high-priority DNNs, boosts overall system throughput, and accounts for resource allocation requirements for all DNNs.
\end{inparaenum}

\subsection{Layer representation}\label{subsec:method:representation}

We first define the input tensor \(\mathcal{Q}\) to represent a multi-DNN mapping. Each channel \(\mathbf{c}^{i}\) in \(\mathcal{Q}\) corresponds to the \(i^{th}\) DNN in the workload. Each row \(j\) in \(\mathbf{c}^{i}\) represents a layer \(\mathbf{l}_{j}^{i}\). We divide each channel \(\mathbf{c}^i\) into \(\mathbf{d}\) column blocks, where each block represents a different computing component. To capture the complexity and size of each layer, we use the 22-dimensional vector formulated in Equation~\ref{eq:dnn_decomposition}. Tensors $\mathbf{ifm}$, $\mathbf{ofm}$, and $\mathbf{w}$ comprise $4$ elements:
\begin{inparaenum}[(\bgroup\bfseries i\egroup)]
    \item the minibatch size;
    \item the number of channels;
    \item the height of the feature map; and
    \item the width of the feature map.
\end{inparaenum}
Finally, $\mathbf{ps}$ is a $6$-dimensional tensor representing the layer's pad-stride information.

{
  \setlength{\abovedisplayskip}{1pt}
  \setlength{\belowdisplayskip}{2pt}
  \small
\begin{equation}\label{eq:dnn_decomposition}
    \mathbf{l}_{j}^{i} = 
    \hspace*{-4em}\begin{tikzpicture}[baseline=(m.center)]
        \matrix (m) [matrix of math nodes, left delimiter=(, right delimiter=), row sep=0.5em, column sep=0.5em] {
            j & \mathbf{t} & \mathbf{ifm} & \mathbf{ofm} & \mathbf{w} & \mathbf{b} & \mathbf{a} & \mathbf{ps} \\
        };
        \draw[-Stealth] (m-1-1.south) -- +(0, -0.65) -- +(-0.5, -0.65) node[left, align=center] {Layer index \\ of DNN $i$};
        \draw[-Stealth] (m-1-3.south) -- +(0, -0.15) node[below, align=center] {Input \\ feature map};
        \draw[-Stealth] (m-1-5.south) -- +(0, -0.15) node[below, align=center] {Weights \\ tensor};
        \draw[-Stealth] (m-1-7.south) -- +(0, -0.65) -- +(0.5, -0.65) node[right, align=center] {Type of \\ activation};
        \draw[-Stealth] (m-1-2.north) -- +(0, 0.65) -- +(-0.5, 0.65) node[left, align=center] {Layer \\ type};
        \draw[-Stealth] (m-1-4.north) -- +(0, 0.15) node[above, align=center] {Output \\ feature map};
        \draw[-Stealth] (m-1-6.north) -- +(0, 0.15) node[above, align=center] {Number of \\ biases};
        \draw[-Stealth] (m-1-8.north) -- +(0, 0.85) -- +(0.2, 0.85) node[right, align=center] {Pad-stride \\ information};
    \end{tikzpicture}\hspace*{-3em}
\end{equation}}

\subsection{Static and Dynamic Prioritization}\label{subsec:method:static-vs-dynamic}
We employ two distinct approaches within the RankMap framework to prioritize DNNs, effectively addressing different operational needs and scenarios. The static priority method is designed for scenarios where a specific, critical DNN is prioritized above others within the workload. This DNN is assigned a high priority $\mathbf{p}^{i}$. This ensures that resources are allocated preferentially to this DNN to meet specific performance requirements. 

In contrast to static priority, dynamic priority adjusts the importance of each DNN based on its computational demands as characterized by the layer profile $\mathbf{l}_{j}^{i}$. This method uses a priority vector that changes dynamically, facilitating more balanced resource distribution across all DNNs in the workload. This flexibility results in higher overall system throughput compared to the static method (Section~\ref{sec:experimental}), as it allows the system to adapt resource allocation based on real-time computational needs. Both priority types are designed to prevent application starvation. To ensure no DNN is starved, RankMap employs a disqualification mechanism, as detailed in Section~\ref{subsec:method:mcts}. Any mapping likely to result in starvation is automatically excluded, thus safeguarding against performance degradation and ensuring efficient resource distribution.

\subsection{Vector Quantized-Variational AutoEncoder}\label{subsec:method:vqvae}

Next, we utilize a Vector Quantized Variational AutoEncoder (VQ-VAE) model~\cite{van2017neural} to effectively compress and encode the raw layer vector representations. The VQ-VAE includes a bottleneck layer that, unlike typical AutoEncoders, introduces a controlled level of stochasticity. It quantizes the data distribution using discrete latent variables, enabling the creation of a highly scalable codebook. Each layer is compressed into a $16$-dimensional distributed embedding~\cite{mikolov2013distributed}, reducing the computational load of our throughput estimator by $\sim 58 \%$ in multiply-accumulate (MAC) operations. To convert the raw layer vectors into latent vectors, we utilize 1D convolutional layers. We apply quantization to the distribution in the latent space using Grouped Residual Vector Quantization~\cite{yang2023hifi}.

\subsection{Throughput Estimator}\label{subsec:method:estimator}

Building on the formulation of the input tensor \(\mathcal{Q}\), we evaluate any mapping \(\mathcal{M}\) using a specialized, multi-task, attention-based convolutional neural network (CNN). This lightweight CNN has approximately $3,7$M parameters and is designed to predict the throughput of each DNN in the workload, expressed in inferences per second, for any given mapping. We approach the throughput estimation for each DNN as an individual task. The multi-task nature of our model allows it to focus on the unique characteristics of each network.

The performance estimator incorporates three shared residual layers that build the backbone and one decoder stream per DNN in the multi-DNN workload. The residual layers are a stack of: 
\begin{inparaenum}[(\bgroup\bfseries i\egroup)]
    \item $\times 2$ depth-wise $2$D convolutional layers and self-attention~\cite{vaswani2017attention} modules; and
    \item a $2$D convolutional layer followed by batch normalization.
\end{inparaenum}
Contrary to standard $2$D convolutions where channels usually exhibit spatial correlations, the channels in \(\mathcal{Q}\) are statistically independent. This necessitates depth-wise convolutions and self-attention modules. Finally, the decoder streams comprise a linear attention layer~\cite{shen2021efficient}, followed by two fully connected layers. Since our performance estimator solves a multi-task problem, linear attention efficiently attends to the feature subset of the encoder concerning each task.

\subsection{Monte-Carlo Tree Search}\label{subsec:method:mcts}

\begin{figure}
    \centering\hspace*{-5pt}
    \resizebox{1.025\columnwidth}{!}{\includegraphics[page=2, width=\linewidth, clip, trim={2em 4 4 3}]{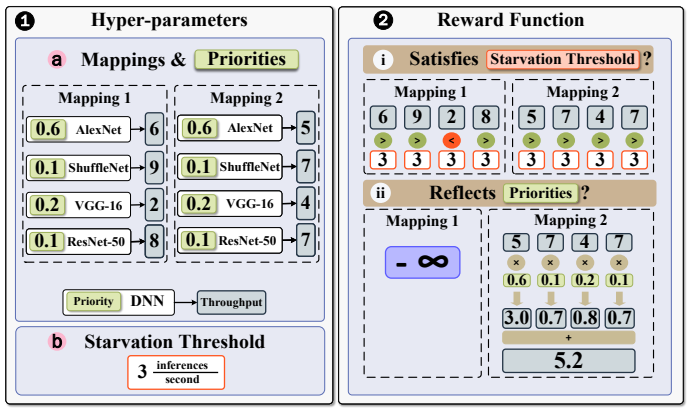}}
    \caption{Reward calculation in RankMap by example.}
    \label{fig:method:rew-calc}
\end{figure}

While the throughput estimator predicts a DNN's performance, finding the optimal mapping on the platform is \textit{challenging due to the vast solution space}. Consider the following multi-DNN workload as an example:
\begin{inparaenum}[(\bgroup\bfseries i\egroup)]
    \item AlexNet;
    \item MobileNet;
    \item ResNet-50; and
    \item ShuffleNet.
\end{inparaenum}
For this combination alone, the total number of possible mappings is calculated as $3^{(8 + 20 + 18 + 18)} \approx 4\mathrm{E}{+10}$. Here, the numbers $\{ 8, 20, 18, 18 \}$ represent the valid partition points for each DNN, and $3$ corresponds to the number of computing components (i.e., GPU, big CPU cluster, and LITTLE CPU cluster). Evaluating such a vast number of mappings is impractical. 

To address this, we use Monte-Carlo Tree Search (MCTS)~\cite{mcts-recent}, which stochastically prunes the search space through a decision tree. MCTS iteratively selects and expands tree nodes based on their rewards, with each node representing a potential mapping. During expansion, a random path under a selected node is explored, using the trained performance estimator to simulate these paths. The estimator's output updates the decision tree's weights, which, combined with a predefined computational budget, prunes the search tree for fast and accurate results. The final output is the mapping with the highest expected reward.

To factor in DNN priorities, we use the priority vector \(\mathbf{p}\) to calculate the reward for a given mapping \(\mathcal{M}\). The throughput estimator's output \(\mathbf{O} (\mathcal{M})\) is multiplied by \(\mathbf{p}\) to generate the reward. \textit{If any element in \(\mathbf{O} (\mathcal{M})\) falls below a specific threshold \(\mathbf{th}\), the state is disqualified from the solution space}. Overall, MCTS solves for the optimal mapping \(\mathcal{M}^{*}\):
{
  \setlength{\abovedisplayskip}{3pt}
  \setlength{\belowdisplayskip}{3pt}
\[
    \mathcal{M}^{*} = \argmax_{\mathcal{M}} \left( \mathbf{O}(\mathcal{M})^{\mathbf{T}} \times \mathbf{p} \; \middle|\; \mathbf{O}(\mathcal{M})^{\mathbf{i}} > \mathbf{th} \;\; \forall \mathbf{i} \in \mathbf{O}(\mathcal{M}) \right)
\]
}which favors mappings meeting $\mathbf{th}$, with their reward reflecting the weighted priorities of the DNNs.

Figure~\ref{fig:method:rew-calc} shows an example of the reward calculation process for two mappings. We assume a threshold $\mathbf{th} = 3$ for our example and a priority vector $[0.6, 0.1, 0.2, 0.1]$ for four DNNs. After obtaining the throughput predictions $\mathbf{O}(\mathcal{M}_i)$ for each mapping from our estimator, we \circled{1} check if any DNN's predicted throughput falls below $\mathbf{th}$. We disqualify mapping $1$ as VGG-16's throughput is below the $\mathbf{th}$; \circled{2} calculate the reward for each mapping:
\begin{inparaenum}[(\bgroup\bfseries i\egroup)]
\item \textbf{Mapping $\mathcal{M}_1$:} Large negative reward (fails $\mathbf{th}$)
\item \textbf{Mapping $\mathcal{M}_2$:} Calculated weighted sum (meets $\mathbf{th}$ for all DNNs)
\end{inparaenum}
This favors mappings meeting $\mathbf{th}$, with their reward reflecting the weighted priorities of the DNNs.

\section{Experimental Evaluation}\label{sec:experimental}

In this section, we demonstrate the efficiency of RankMap in terms of: 
\begin{inparaenum}[(\bgroup\bfseries i\egroup)]
  \item increased average system throughput  and enhanced performance for high-priority DNNs (Section~\ref{subsec:experimental:throughput-prioritization});
  \item prevention of starvation (Section~\ref{subsec:experimental:starvation});
  \item correlation between throughput and priorities (Section~\ref{subsec:experimental:corr}); and 
  \item run-time trade-off (Section~\ref{subsec:experimental:runtime}).
\end{inparaenum}
We create several diverse multi-DNN workloads on the Orange Pi 5 board, a heterogeneous embedded platform featuring a Mali-G610 GPU and big.LITTLE CPUs with quad-core Cortex-A76 and quad-core Cortex-A55 running at 2.4GHz and 1.8GHz, respectively. RankMap is implemented in PyTorch. ARM Compute Library~\cite{armcl} and OpenCL are used to execute DNNs on the board and for DNN partitioning.

To train our multi-task throughput estimator, we created a dataset of $10$K workloads. Each workload consists of a mix of up to 5 concurrent DNNs randomly selected from a pool of 23 DNNs. The DNNs in the pool are:
\begin{inparaenum}[(\bgroup\bfseries i\egroup)]
  \item AlexNet,
  \item DenseNet-121,
  \item DenseNet-169,
  \item EfficientNet-B0,
  \item EfficientNet-B1,
  \item EfficientNet-B2,
  \item GoogleNet,
  \item Inception-ResNet V2,
  \item Inception V3,
  \item Inception V4,
  \item MobileNet,
  \item MobileNet V2,
  \item ResNet-12,
  \item ResNet-50,
  \item ResNet-50 V2,
  \item ResNeXt-50,
  \item ShuffleNet,
  \item SqueezeNet,
  \item SqueezeNet V2,
  \item SSD with MobileNet backbone,
  \item YOLO V3,
  \item VGG-16, and
  \item VGG-19.
\end{inparaenum}

We randomly partitioned each DNN and mapped the sub-DNNs across the device's computing components, creating a diverse and sizable dataset. This randomness ensures the dataset represents the entire solution space, with each sample being \ul{unique}. We executed each workload on the board, recording the inferences per second (throughput) for each DNN. We collected $10$K samples, using $90\%$ for training our throughput estimator. Although the estimator will be tested on real-world \ul{unseen} scenarios during experiments, we reserved $10\%$ of the dataset for feedback during training. Using L2-loss for each decoder stream, our estimator achieved an L2 loss of about $0.14$ after $50$ epochs. Random channel shuffling as a dataset augmentation step further reduced the L2 loss to about $0.08$.

We consider the following metrics:
\begin{inparaenum}[(\bgroup\bfseries i\egroup)]
  \item \textbf{Normalized throughput \(\mathcal{T}\)}: The system throughput is calculated as \(\mathcal{T} = \frac{\sum^{N}_{i=1}t_{current}^{i}}{N}\), where \(t_{current}\) represents the throughput in inferences per second for each DNN $i$ when operating concurrently within the workload. This total throughput is then normalized against the baseline, i.e., the scenario where all DNNs are executed on the GPU, the highest-performing component on the Orange Pi 5; and
  \item \textbf{Potential throughput \(\mathcal{P}\)}: Defined as \(\mathcal{P} = \frac{t_{current}^{i}}{t_{ideal}^{i}}\), where $t_{ideal}$ is the throughput of the DNN when executed alone on the GPU. This metric assesses how the performance of each DNN in a shared environment compares to its performance in isolation. Potential throughput provides a quantitative measure to see how close a DNN's current performance is relative to its optimal performance.
\end{inparaenum}

Additionally, in the following experiments, we evaluated RankMap against:
\begin{inparaenum}[(\bgroup\bfseries i\egroup)]
  \item \textbf{Baseline};
  \item \textbf{MOSAIC}~\cite{han2019mosaic}, a linear regression approach;
  \item \textbf{ODMDEF}~\cite{lim2021odmdef}, a manager with both a linear regression and a $k$-NN classifier in its core;
  \item \textbf{GA}, the evolutionary manager proposed in~\cite{kang2020scheduling}; and
  \item \textbf{OmniBoost}~\cite{karatzas2023omniboost}, a framework for greedy throughput optimization of multi-DNN workloads.
\end{inparaenum}
Regarding RankMap, we considered:
\begin{inparaenum}[(\bgroup\bfseries i\egroup)]
  \item \textbf{Static mode} RankMap$_\mathcal{S}$, where a high priority is assigned to a critical DNN operating concurrently with others; and
  \item \textbf{Dynamic mode} RankMap$_\mathcal{D}$, where the priority vector for each DNN is derived from its profiling stage, as detailed in Section~\ref{subsec:method:static-vs-dynamic}.
\end{inparaenum}

\subsection{Throughput and Prioritization comparison}\label{subsec:experimental:throughput-prioritization}

To evaluate the efficiency of RankMap in terms of throughput and prioritization, we created multi-DNN workloads with 3, 4, and 5 DNNs operating concurrently. For each set of workloads, we evaluated both the normalized throughput \(\mathcal{T}\) and the potential throughput \(\mathcal{P}\). First, we focus on the potential throughput of the DNN with the \textit{highest} priority to highlight how well each manager evaluates key applications and supports the critical DNN in a workload (extensive analysis of all DNNs' performance with respect to their assigned priority to illustrate RankMap's capabilities in managing diverse DNN workloads follows in Section~\ref{subsec:experimental:corr}).

\begin{figure}
\centering
  \resizebox{1.0\columnwidth}{!}{\includegraphics[page=1, width=\linewidth, clip]{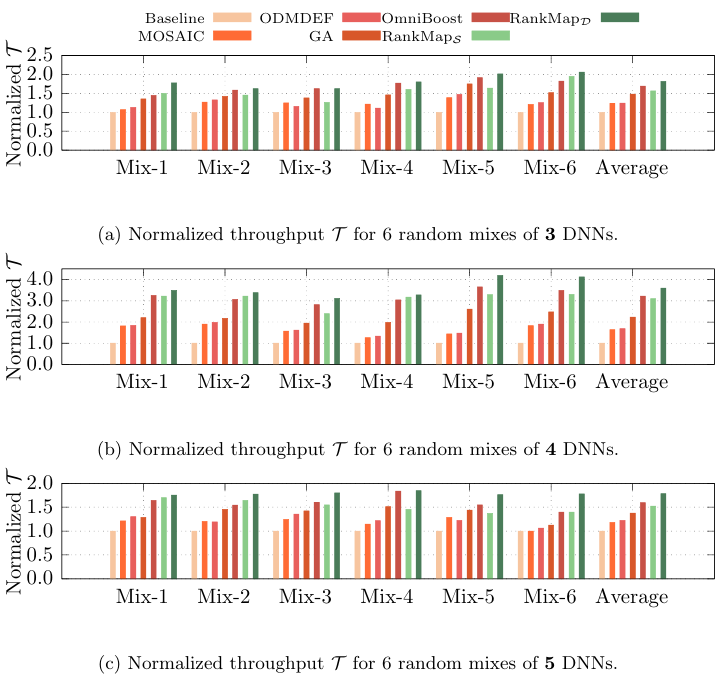}}
  \caption{Normalized throughput $\mathcal{T}$ across random mixes of $\mathbf{3}$, $\mathbf{4}$, and $\mathbf{5}$ concurrent DNNs.}
  \label{fig:mixU:throughput}
\end{figure}

\begin{figure}
    \centering
    \resizebox{1.0\columnwidth}{!}{\includegraphics[page=1, width=\linewidth, clip]{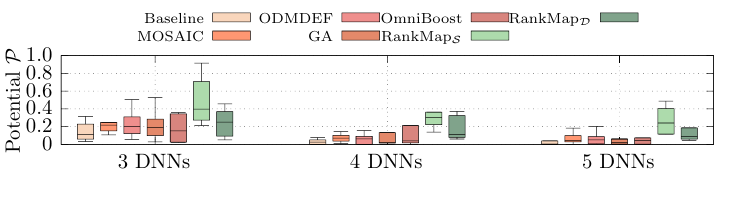}}
    \caption{Potential throughput $\mathcal{P}$ of the high-priority DNN across mixes of $\mathbf{3}$, $\mathbf{4}$, and $\mathbf{5}$ concurrent DNNs.}
    \label{fig:mixU:priority}
\end{figure}

\begin{figure*}
  \centering
  \resizebox{1.0\textwidth}{!}{\includegraphics[page=1, width=\linewidth, clip]{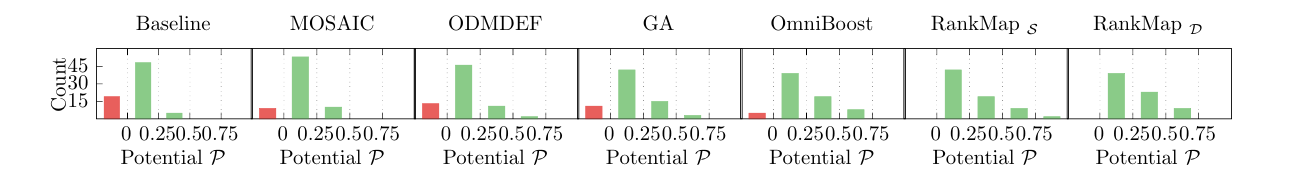}}
  \caption{Comparison of Potential Throughput $\mathcal{P}$ across all the performed experiments.}
  \label{fig:starvation}
\end{figure*}

\textbf{Mixes of 3 DNNs:} Figure~\ref{fig:mixU:throughput} shows that RankMap$_\mathcal{D}$ achieves on average $82 \%$, $48 \%$, $46 \%$, $23 \%$, and $8 \%$ higher $\mathcal{T}$ than the Baseline, MOSAIC, ODMDEF, GA, and OmniBoost, respectively. The RankMap$_\mathcal{S}$ falls behind the dynamic by $15 \%$. However, RankMap$_\mathcal{S}$ keeps the potential throughput $\mathcal{P}$ higher by $\times 6.3$, $\times 2.2$, $\times 2.7$, $\times 4$, and $\times 2.6$ than the Baseline, MOSAIC, ODMDEF, GA, OmniBoost, and RankMap$_\mathcal{D}$, respectively (Figure~\ref{fig:mixU:priority}). Notably, RankMap$_\mathcal{S}$ maintains higher $\mathcal{P}$ than $0.21$ under any workload, with a peak value of $0.91$.

\textbf{Mixes of 4 DNNs:} Regarding $4$ concurrent DNNs, we observe that some managers have already overwhelmed the GPU. Specifically, Figure~\ref{fig:mixU:throughput} depicts that RankMap$_\mathcal{D}$ demonstrated $\times 3.6$, $\times 2.2$, $\times 2.1$, $\times 1.6$, and $\times 1.2$ higher normalized throughput $\mathcal{T}$ compared to the Baseline, MOSAIC, ODMDEF, GA, and OmniBoost, respectively. RankMap$_\mathcal{S}$ fell behind by $14 \%$ in terms of $\mathcal{T}$ compared to RankMap$_\mathcal{D}$. However, Figure~\ref{fig:mixU:priority} justifies this trade-off since RankMap$_\mathcal{S}$ demonstrated higher $\mathcal{P}$ by $\times 57.5$, $\times 7.4$, $\times 35.1$, $\times 21.9$, and $\times 2.2$  than the Baseline, MOSAIC, ODMDEF, GA, OmniBoost, and RankMap$_\mathcal{D}$, respectively. Notably, RankMap$_\mathcal{S}$ keeps $\mathcal{P}$ higher than $0.14$ under any workload, with a peak value of $0.37$. Hence, under heavy workload, RankMap$_\mathcal{S}$ satisfies the critical DNN's performance requirements while maintaining high system throughput.

\textbf{Mixes of 5 DNNs:} To further stress the device's resources, we evaluate the managers for 5 concurrent DNNs (Figures~\ref{fig:mixU:throughput} and~\ref{fig:mixU:priority}). While under heavy workload, RankMap$_\mathcal{D}$ still finds solutions that increase the average system throughput by $79 \%$, $51 \%$, $46 \%$, $30 \%$, and $12 \%$ compared to the Baseline, MOSAIC, ODMDEF, GA, and OmniBoost, respectively. RankMap$_\mathcal{S}$ falls again third behind RankMap$_\mathcal{D}$ and OmniBoost by $5 \%$ and $17 \%$, respectively. Still, it yields $\times 55$, $\times 12$, $\times 18$, $\times 42$, and $\times 38$ higher $\mathcal{P}$ than the Baseline, MOSAIC, ODMDEF, GA, and OmniBoost, respectively. Furthermore, RankMap$_\mathcal{S}$ keeps $\mathcal{P}$ higher than $0.12$ under any workload, with a peak value of $0.49$. Hence, RankMap$_\mathcal{S}$ effectively prioritizes DNNs even under heavy workloads, while also boosting system throughput.

\subsection{Starvation comparison}\label{subsec:experimental:starvation}

Next, we determine if there were any instances of DNN starvation within the workload mixes discussed in Section~\ref{subsec:experimental:throughput-prioritization}. To that end, we analyzed the occurrence of starved DNNs across the $72$ samples from these mixes, as illustrated in Figure~\ref{fig:starvation}. We define a DNN as starved when its potential throughput ($\mathcal{P}$) is 0, represented by the red bin in Figure~\ref{fig:starvation}. Our findings show that both the RankMap$_\mathcal{S}$ and RankMap$_\mathcal{D}$ successfully prevent DNN starvation under all tested workloads. In contrast, Baseline, MOSAIC, ODMDEF, GA, and OmniBoost returned mappings that starved some DNNs, with respective counts of $19$, $9$, $13$, $11$, and $5$ starved DNNs out of $72$. Therefore, in pursuit of higher throughput, some managers caused starvation effects by throttling certain DNNs.

\begin{figure}
  \centering
  \resizebox{1.0\columnwidth}{!}{\includegraphics[page=1, width=\linewidth, clip]{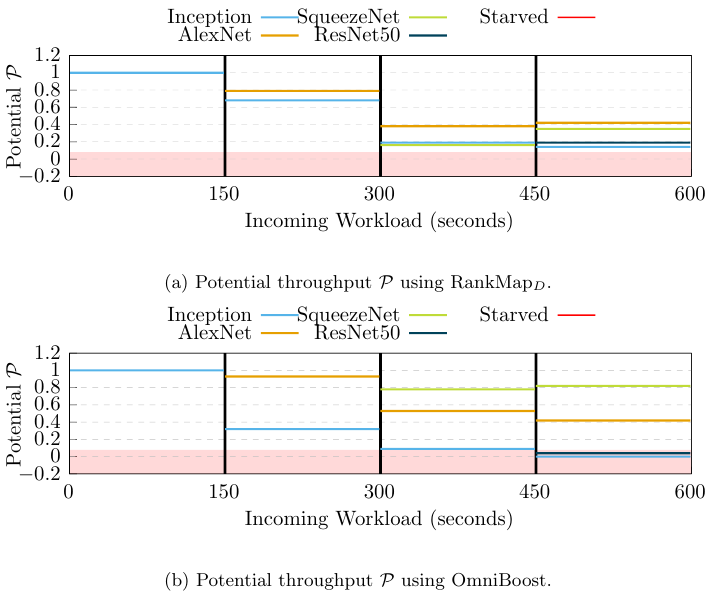}}
  \caption{Comparison on a $4$-DNN dynamic workload.}
  \label{fig:dwork}
\end{figure}

To further demonstrate this behavior, consider a dynamic workload comprising four DNNs, each with a unique computational profile, arriving at different times. The DNNs and their order of arrival are as follows:
\begin{inparaenum}[(\bgroup\bfseries i\egroup)]
  \item \textit{Inception ResNet-V1}, a highly demanding model ($t_{ideal} = 4 $ inferences per second);
  \item \textit{AlexNet}, a model with standard computational demands ($t_{ideal} = 43 $ inferences per second);
  \item \textit{SqueezeNet-V1}, a lightweight model ($t_{ideal} = 67 $ inferences per second); and
  \item \textit{ResNet-50}, a demanding model ($t_{ideal} = 20 $ inferences per second);
\end{inparaenum}
According to the analysis in Section~\ref{subsec:experimental:throughput-prioritization}, RankMap$_{D}$ achieved
the highest throughput $\mathcal{T}$ across all experiments, with OmniBoost being the next best one.
To that end, we compare RankMap$_{D}$ with OmniBoost in Figure~\ref{fig:dwork} regarding $\mathcal{P}$ during this multi-DNN workload.
\begin{itemize}[leftmargin=*]
  \item Initially, Inception arrives first and both RankMap$_{D}$ and OmniBoost place it on the GPU achieving the ideal throughput.
  \item At $time=150$, AlexNet arrives. Given that Inception is more computationally intensive, RankMap$_{D}$ assigns higher priority to it, maintaining a high potential throughput $\mathcal{P}$. In contrast, OmniBoost aims for a mapping that favors AlexNet, believing it will result in a higher average throughput as it is more lightweight. Consequently, the potential throughput of Inception significantly decreases.
  \item At $time=300$, SqueezeNet arrives. RankMap$_{D}$ continues to prioritize Inception, trying to find a mapping that supports it efficiently. OmniBoost, however, largely neglects Inception, focusing instead on the other two networks which it perceives to contribute more to the average system throughput. 
  \item At $time=450$ when ResNet-50 arrives, the system becomes oversubscribed. OmniBoost results in starvation for both Inception and ResNet, whereas RankMap$_{D}$ manages to avoid starvation for any DNN, ensuring even the more demanding networks progress.
\end{itemize}
Comparing only the average system throughput, OmniBoost is better achieving $\mathcal{T} = 18$, while RankMap$_{D}$ achieves $\mathcal{T} = 14$. \emph{However, while OmniBoost strives to maximize throughput, it ends up throttling two DNNs in the workload, causing significant starvation effects.}

\begin{figure}[h]
  \centering
  \resizebox{1.0\columnwidth}{!}{\includegraphics[page=1, width=\linewidth, clip]{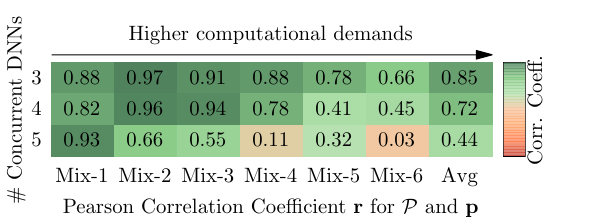}}
  \caption{Pearson corr. coef. of Potential Throughput $\mathcal{P}$ and priority vectors $\mathbf{p}$ for the mixes of 3, 4, and 5 concurrent DNNs for RankMap$_\mathcal{D}$.}
  \label{fig:exp:pearson-dyn}
\end{figure}

\subsection{Throughput and Priority correlation}\label{subsec:experimental:corr}

The previous analysis primarily focused on the high-priority DNN. Therefore, we still need to evaluate how effectively RankMap satisfies the assigned priorities for all DNNs in the mix. As mentioned earlier, RankMap$_\mathcal{D}$ assigns priorities to the DNNs based on their computational profiles. Thus, for RankMap$_\mathcal{D}$, we compiled a heatmap in Figure~\ref{fig:exp:pearson-dyn} that depicts how well potential throughput $\mathcal{P}$ correlates with the assigned priority vectors $\mathbf{p}$. The mixes depicted are directly ported from Section~\ref{subsec:experimental:throughput-prioritization}. We also sorted workloads from least to most computationally demanding ones. We employ the Pearson correlation coefficient $\mathbf{r} \in [-1, 1]$ to quantify the relationship between $\mathcal{P}$ and $\mathbf{p}$. 

In Figure~\ref{fig:exp:pearson-dyn}, we observe that for mixes of $3$ concurrently executing DNNs, RankMap$_\mathcal{D}$ yields mappings that highly satisfy the provided priority vectors $\mathbf{p}$, resulting in $0.85$ average $\mathbf{r}$. Naturally, RankMap$_\mathcal{D}$ has more difficulty satisfying the priorities assigned as the computational demands of the workloads increase. This is also evident with mixes of $4$ concurrently executing DNNs, where Mix-$5$ and Mix-$6$ score below $0.5$ $\mathbf{r}$. This is because all devices begin to saturate, and for RankMap$_\mathcal{D}$ to prevent starvation for all DNNs, it deviates from the given priority vector. Finally, the platform saturates completely in several mixes of $5$ DNNs. Notably, even under these extreme conditions, RankMap$_\mathcal{D}$ maintains positive $\mathbf{r}$ values. This indicates that our framework continues to strive for a balanced mapping that considers throughput, prioritization constraints, and starvation prevention, even in highly resource-constrained settings.

\begin{figure}[h]
  \centering
  \resizebox{1.0\columnwidth}{!}{\includegraphics[page=1, width=\linewidth, clip]{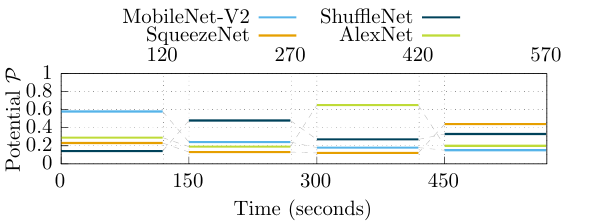}}
  \caption{Potential Throughput $\mathcal{P}$ for different priority vectors $\mathbf{p}$ using RankMap$_\mathcal{S}$ for the mix of: \textbf{(i)} MobileNet-V2; \textbf{(ii)} SqueezeNet-V1; \textbf{(iii)} ShuffleNet; and \textbf{(iv)} AlexNet.}
  \label{fig:exp:runtime-priority-turnaround}
\end{figure}

RankMap$_\mathcal{S}$ operates based on user-defined priorities. To demonstrate its efficacy in balancing throughput optimization while satisfying priority constraints and preventing throttling effects across all DNNs in a workload, we present a scenario illustrated in Figure~\ref{fig:exp:runtime-priority-turnaround}. This scenario comprises a workload of four concurrent DNNs:
\begin{inparaenum}[(\bgroup\bfseries i\egroup)]
  \item MobileNet-V2; 
  \item ShuffleNet;
  \item AlexNet; and
  \item SqueezeNet.  
\end{inparaenum}
The experiment shows dynamic changes in user priorities across multiple stages. In the initial stage, the user-defined priority vector $\mathbf{p}{1}$ is set to $(0.7, 0.1, 0.1, 0.1)$, with RankMap$_\mathcal{S}$ constructing a mapping that adheres to these priorities. Then, in stage two, the user changes the priorities to $\mathbf{p}{2} = (0.1, 0.7, 0.1, 0.1)$, shifting the highest priority to ShuffleNet, a less computationally demanding network compared to MobileNet-V2. RankMap$_\mathcal{S}$ adapts to this change, satisfying the new priority configuration while ensuring no DNN experiences starvation. This adaptive behavior persists through the remaining stages of priority shifts to the other two DNNs. The dashed grey lines indicate the time it takes for RankMap$_\mathcal{S}$ to identify the appropriate mappings, detailed further in Section~\ref{subsec:experimental:runtime}.

\subsection{Run-time performance evaluation}\label{subsec:experimental:runtime}

The Baseline manager is the fastest in response time as it directly maps all DNNs to the GPU, but fails to utilize the platform's heterogeneity, leading to suboptimal throughput, prioritization, and increased starvation risk. MOSAIC, slightly faster than ODMDEF (both around 1 second), quickly causes device saturation as demonstrated in sections~\ref{subsec:experimental:throughput-prioritization} and~\ref{subsec:experimental:starvation}, resulting in low throughput and starvation, which shows the limitations of simplistic designs. In contrast, the GA is the slowest, requiring evaluations for each chromosome on the Orange Pi 5 for every generation, and it cannot use past data to predict or adapt to new workloads. OmniBoost and our RankMap both show a run-time of about 30 seconds, suggesting that RankMap achieves an optimal balance of run-time performance and efficient workload management, supporting high throughput, prioritization, and starvation prevention.

\section{Conclusion}\label{sec:conclusion}

Modern application workloads, involving multiple concurrent DNNs, pose significant challenges for heterogeneous embedded systems focused on optimizing throughput and managing priorities without causing starvation. This paper presents RankMap, a scalable, efficient manager that enhances system performance. RankMap boosts average throughput by up to $\times 3.6$ and improves resource allocation effectiveness by up to $\times 57.5$ compared to existing solutions. Importantly, it prevents application starvation across multi-DNN workloads.

\section*{Acknowledgments}

This work is supported by grant NSF CCF 2324854.

\bibliographystyle{IEEEtran}
\footnotesize
\bibliography{ref}
\end{document}